\documentclass{article}

     \PassOptionsToPackage{round}{natbib}
     \usepackage[preprint]{neurips_2022}

\usepackage[utf8]{inputenc} % allow utf-8 input
\usepackage[T1]{fontenc}    % use 8-bit T1 fonts
\usepackage{hyperref}       % hyperlinks
\usepackage{url}            % simple URL typesetting
\usepackage{booktabs}       % professional-quality tables
\usepackage{amsfonts}       % blackboard math symbols
\usepackage{nicefrac}       % compact symbols for 1/2, etc.
\usepackage{microtype}      % microtypography
\usepackage{xcolor}         % colors

\usepackage{nicefrac}       % compact symbols for 1/2, etc.
\usepackage{microtype}      % microtypography
\usepackage{subfig}         % subfig
\usepackage{graphicx}       % multiple figures
\usepackage{amsmath}        % equation
\usepackage{amsmath,amssymb,amsfonts}
\usepackage{threeparttable} % For multrow tables
\usepackage{multirow} % For multrow tables
\usepackage[ruled, lined, linesnumbered, commentsnumbered, longend]{algorithm2e}

\title{CTMQ: Cyclic Training of Convolutional Neural Networks with Multiple Quantization Steps}

\author{%
  HyunJin Kim\thanks{Corresponding Author, homepapge: https://www.empaslab.com} \\
  School of Electronics and Electrical Engineering, \\
  Dankook University\\
  152, Jukjeon-ro, Suji-gu, Yongin-si, Gyeonggi-do, Republic of Korea\\
  \texttt{hyunjin2.kim@gmail.com} \\
   \And
  Jungwoo Shin\\
  School of Electronics and Electrical Engineering, \\
  Dankook University\\
  152, Jukjeon-ro, Suji-gu, Yongin-si, Gyeonggi-do, Republic of Korea\\
  \texttt{sjo506@naver.com} \\
  \AND
  Alberto A. Del Barrio \\
  Department of Computer Architecture and System Engineering, Complutense University of Madrid, Madrid, ES \\
  \texttt{abarriog@ucm.es} \\
}

\begin{document}

\maketitle

\begin{abstract}
This paper proposes a training method having multiple cyclic training for achieving enhanced performance 
in low-bit quantized convolutional neural networks (CNNs). 
Quantization is a popular method for obtaining lightweight CNNs, 
where the initialization with a pretrained model is widely used 
to overcome degraded performance in low-resolution quantization.
However, large quantization errors between real values and their low-bit quantized ones
cause difficulties in achieving acceptable performance for complex networks and large datasets.
The proposed training method softly delivers the knowledge of pretrained models 
to low-bit quantized models in multiple quantization steps.
In each quantization step, 
the trained weights of a model are used to initialize the weights of the next model 
with the quantization bit depth reduced by one.
With small change of the quantization bit depth,
the performance gap can be bridged, thus providing better weight initialization. 
In cyclic training, 
after training a low-bit quantized model, 
its trained weights are used in the initialization of its accurate model to be trained.
By using better training ability of the accurate model in an iterative manner, 
the proposed method can produce enhanced trained weights for the low-bit quantized model 
in each cycle.   
Notably, the training method can advance Top-1 and Top-5 accuracies of the binarized ResNet-18
on the ImageNet dataset by 5.80\% and 6.85\%, respectively.
\end{abstract}

\section{Introduction}
\label{section:Introduction}

In recent years, quantized CNNs are widely adopted 
for reducing the hardware complexity 
(\citet{gupta2015deep, han2015deep, wu2016quantized, hubara2017quantized, wang2018training, wu2020integer, yu2020low}).
Moreover, it has been proved that low-bit quantized CNNs using binary 
(\citet{hubara2016binarized, courbariaux2016binarized, rastegari2016xnor,liu2018bi}), 
ternary (\citet{zhou2016dorefa, deng2017gated, wan2018tbn}) 
 quantizations can be trainable, producing acceptable performance. 
However, there is a significant performance gap between real-valued CNNs and low-bit quantized CNNs.
Quantization noise disturbs weight updating in the quantization-aware training (\citet{wang2018training}), 
thus degrading the performance of trained models in low-bit quantized CNNs.
The retraining method uses pretrained weights to initialize a target model, 
and the initialized model is retrained. 
Because the retraining method automatically searches for suitable parameter values,
it can produce better classification results 
compared with naive post-training quantization (\citet{gholami2021survey}).
It is known that the initial weights from accurate models are helpful to produce better training results. 
However, low-bit quantized CNNs show significant performance degradation immediately after the initialization.

In each batch of training, after updating real-valued weights in back-propagation, 
quantized weights are used in forward paths, making quantization noise.  
The quantization noise hinders the training step to find its optimized weights (\citet{helwegen2019latent, xu2021improving}). 
Besides, the scaling and biasing parameters used in normalization 
have disturbances due to the gap between real-valued parameters and their quantized values. 
There have been several works to overcome the degraded performance in low-bit quantized models (\citet{helwegen2019latent, kim2020binaryduo, martinez2020training, xu2021improving, xu2021recu}).
However, the existing studies for low-bit quantized models 
do not consider the training method for achieving better weight initialization from accurate models.

This paper proposes a method consisting of multiple cyclic training for achieving better performance 
in low-bit quantized CNNs.
Firstly, the proposed method \textit{softly} transfers the learned knowledge based on accurate models 
to low-bit quantized models with multiple quantization steps.
In a quantization step, 
the pretrained weights are used to initialize the weights of the next model 
with the quantization bit depth reduced by one.
With small change of the quantization bit depth,
the performance gap can be bridged, thus producing better weight initialization. 
In the multiple quantization steps, the pretrained weights in a step 
are used in the initialization of weights to be trained in the next quantization step.
As proceeding with each quantization step, 
the quantization bit depth of trained models decreases gradually. 
Secondly, the proposed cyclic training uses the trained weights of a low-bit quantized model
 in the weight initialization of its accurate model to be trained.
Then, after training the accurate model, 
its trained weights are used in the training of the low-bit quantized model in an iterative manner. 
The proposed method uses better training ability of the accurate model in each cycle,
producing better weight initialization for the low-bit quantized model in an iterative manner.   
The proposed training method is applied to the training of the binarized ResNet-18 (\citet{he2016deep}). 
Quantized CNNs based on DoReFa-Net (\citet{zhou2016dorefa}) and XNOR-Net (\citet{rastegari2016xnor}) 
are trained with the proposed method. 
Without any structural changes in ResNet-18,
the proposed training method outperforms 
the training results from scratch and using the initialization of pretrained real-valued models. 
In the experiments on the ImageNet dataset, 
Top-1 and Top-5 accuracies of the binarized ResNet-18 are enhanced by 5.80\% and 6.85\%, respectively.

This paper is organized as follows: 
in the preliminaries, the difficulties in training highly approximate CNNs are described in detail.
Then, the proposed cyclic training having multiple quantization steps is explained 
along with a detailed process description. 
Next, we explain our experimental environments and experimental results for training the binarized ResNet-18.

\section{Preliminaries}
\label{section:Preliminaries}

\subsection{Quantized CNNs}
As the easiest way to reduce the computational resources and storage requirements of CNNs, 
quantized CNNs have been studied recently.
Normalized activations and weights used in the forward path of CNNs 
are expressed using a fixed-point format.
Quantized CNNs quantize activations and weights to have 8-bit, 4-bit, 2-bit, ternary, and binary formats.
The simplified format and its lightweight operations 
can dramatically reduce the hardware complexity of convolutional operations in CNNs,
compared with real-valued CNNs.
Notably, the binarized CNNs utilize 1-bit activations and weights in binarized convolutions.

It has been proven in several existing studies
that quantized CNNs can achieve a certain level of inference performance (\citet{gholami2021survey}).
However, due to quantization errors, 
their performance can be degraded compared with real-valued CNNs.
Notably, there may be a significant performance degradation
in the quantized CNN using weights 
based on a format of less than 2-bit (\citet{zhou2016dorefa}).
Therefore, it is necessary to improve the performance of quantized CNNs. 

When a pretrained CNN model is given, the pretrained model can be used to initialize its quantized CNN model.
Quantization of CNNs has been categorized depending on whether retrained is performed or not.
Post-training quantization reuses the parameters of the pretrained model without retraining,
where quantization levels are uniformly spaced or unequally determined 
to minimize the effects of quantization errors (\citet{gholami2021survey}). 
However, due to the non-linearity in each layer of the CNN model, 
it is difficult to guarantee the best value by reducing the amount of quantization errors. 
On the other hand, the retraining scheme (\citet{han2015deep}) uses the existing pretrained model for the weight initialization of a model
and performs training on the quantized CNN model.  
In quantization-aware training (\citet{wang2018training}), 
the error of the forward path in the quantized model is considered as the amount of loss.
Besides, the parameters for scaling and normalizing activations can be optimized during retraining.
Therefore, it is known that the performance after retraining can be higher than that without retraining 
in quantized CNN models.

\subsection{Training of Quantized CNNs and its Difficulties}

In general, the objective function of a CNN is a non-convex loss function denoted as $\mathcal{L}$,
where training tries to find a local optimum using the gradient descent method (\citet{ruder2016overview}). 
The gradient descent is described in Algorithm~\ref{algo:gd}.

\begin{algorithm}[t]
 \caption{Gradient Descent}
 \small 
 \label{algo:gd}
  \SetKwFunction{Loss}{$\mathcal{L}$}
  \SetKwFunction{Return}{Return}
  \SetKwInOut{KwIn}{Input}
  \SetKwInOut{KwOut}{Output}
  \KwIn {initial weights $w^{(0)}$, number of iterations $T$, learning rate $\gamma$}
  \KwOut {final weights $w^{(T)}$}
  \For{$t \leftarrow 0$ \KwTo $T - 1$}
  {
    $w^{(t + 1)} := w^{(t)} + \gamma \nabla$ \Loss{$w^{(t)}$}
  }
  \Return{$w^{(T)}$}. 
\end{algorithm}

Although other optimizers such as ADAM (\citet{kingma2014adam}) have their own formulations, 
the gradient descent was explained using (\ref{algo:gd}) for its simplicity. 
The local optimum is repeatedly searched by updating the gradient of $\mathcal{L}$. 
The updated weights $w^{(t + 1)}$ are considered when obtaining the next $\mathcal{L}$ in an iteration.

On the other hand, in quantized CNNs, 
the quantization error could nullify or suppress the effects of the updating 
with $\gamma \nabla $ \Loss{$w^{(t)}$} in Algorithm~\ref{algo:gd}.
Weights $w^{(t)}$ at the $t$-th iteration can be rewritten using $k$-bit quantized weights 
$w^{(t)}_{q_k}$ and their following error $\epsilon(w^{(t)}_{q_k})$ as: 

\begin{equation}\label{eq:w_error}
   w^{(j)}_{q_k} := w^{(j)} - \epsilon(w^{(j)}_{q_k}), 0 \leq t \leq T.
\end{equation}

From (\ref{eq:w_error}), $\epsilon(w^{(t)}_{q_k})$ and 
$\epsilon(w^{(t+1)}_{q_k})$ affect the quantized weights used in the next forward path.
The equation in line 2 of Algorithm~\ref{algo:gd} can be rewritten as: 

\begin{equation}\label{eq:gd_error}
   w^{(t + 1)}_{q_k} := w^{(t)}_{q_k} + \epsilon(w^{(t)}_{q_k})
     + \gamma \nabla \mathcal{L}(w^{(t)}_{q_k}) - \epsilon(w^{(t+1)}_{q_k}).
\end{equation}

From (\ref{eq:gd_error}), we can conclude two facts: firstly, 
terms $\epsilon(w^{(t)}_{q_k})$ and $-\epsilon(w^{(t+1)}_{q_k})$ have different signs,
which could somewhat cancel out the quantization error of the updated weights $\epsilon(w^{(t+1)}_{q_k})$.

Secondly, entire values of $\gamma \nabla \mathcal{L}(w{(t)}^{q_k})$ in (\ref{eq:gd_error}), 
cannot be considered in the weight updating 
when $\epsilon(w^{(t)}_{q_k})$ and $-\epsilon(w^{(t+1)}_{q_k})$ are not the same.
When $k$ becomes small in low-bit quantized CNNs, 
the cancellation can be hard, which can mislead searches for the local optimum
in training steps.
It is assured that when a model has high quantization bit depth with small $\epsilon(w^{(t+1)}_{q_k})$,
it can produce good training results.

Several training methods for quantized CNNs empirically adopt a straight-through estimator (STE) (\citet{bengio2013estimating, yin2019understanding})
for back-propagating gradients instead of using derivatives of the quantizer. 
For example, binarized CNNs can use the $sign$ function to binarize activations (\citet{zhou2016dorefa, rastegari2016xnor, liu2018bi}). 
The derivative of the $sign$ function can be the Dirac Delta function (\citet{hassani2009dirac}), 
which is hard to implement. 
Using a STE, the approximation of the derivative of the $sign$ function can be shown by (\ref{eq:ste_sign}).

\begin{equation}\label{eq:ste_sign}
 \nabla sign(x) \approx \left\{\begin{matrix}
 & +1 & \textrm{if }  |x|  \leq 1, \\ 
 & 0 & \textrm{else.} 
\end{matrix}\right.
\end{equation}

Instead of using the Dirac Delta function, low-bit quantized CNNs adopt the STE, making the incoming gradients 
equal to their output gradients in the back-propagation. 

In spite of long training time, 
a STE in low-bit quantized and binarized CNNs can produce acceptable performance in many several works 
(\citet{zhou2016dorefa, rastegari2016xnor, liu2018bi, kim2021aresb, shin2022presb}). 
However, term $\epsilon(w^{(t)}_{q_k})$ in (\ref{eq:w_error}) still exists, 
producing instability in training steps.

\section{CTMQ: cyclic training with multiple quantization steps}
\label{section:CTMQ}

\subsection{Proposed training method}

As discussed in the preliminaries section, 
accurate models with small quantization errors can provide better training performance 
along with more stable weight convergence.
In existing works of binarized CNNs (\citet{liu2018bi, bulat2019xnor, kim2020binaryduo}, 
pretrained weights are used in the weights initialization of the binarized models, 
which can transfer the training results of accurate models into low-bit quantized models.
Let $w(i)^{(T)}$ be the pretrained weights after training an $i$-bit quantized CNN model.
Let us assume that the pretrained weights are used in weight initialization for $k$-bit quantized model. 
The weight updating in (\ref{eq:gd_error}) can be rewritten by $w^{(0)}=w(i)^{(T)}$ as: 
 
\begin{equation}\label{eq:gd_error2}
   w^{(t + 1)}_{q_k} := w(i)^{(T)}_{q_k} + \epsilon(w(i)^{(T)}_{q_k})
     + \gamma \nabla \mathcal{L}(w(i)^{(t)}_{q_k}) - \epsilon(w^{(t+1)}_{q_k}).
\end{equation}

Depending on the difference between $i$ and $k$,
the distribution of quantized values can be different. 
For example, small weights in a Gaussian distribution can be quantized into non-zero values in accurate models. 
On the other hand, when quantization bit depth becomes low in (\ref{eq:gd_error}), 
many small weights are quantized into zeros. 
The significantly different quantization bit depths could prevent the knowledge of pretrained models
from being well transferred into low-bit quantized models.

To overcome the problem, 
we proposed a new method called CTMQ, which is an abbreviation for cyclic training with multiple quantization steps.
In the proposed method, 
the knowledge of a pretrained model is softly transferred into its low-bit quantized model using weight initialization.
Then, the knowledge transfer using initial weights is cyclically repeated 
between $k$ and $(k+1)$-bit quantized models.
In the cyclic training, the pretrained weights of the $k$-bit quantized model 
can be used in the initialization for training the $(k+1)$-bit quantized model in an iterative manner. 
Then, the $(k+1)$-bit quantized model is trained with the initial weights. 
After obtaining the pretrained weights of the $(k+1)$-bit quantized model, 
the pretrained weights can be used the training of the $k$-bit quantized model.
The cyclic training is repeated several times.
In each cycle, 
based on the initial weights from the $k$-bit quantized model, 
the training of the $(k+1)$-bit quantized model 
can produce better training weights for the initialization of the $k$-bit quantized model.
The proposed CTMQ for training a $k$-bit quantized model is described in Algorithm~\ref{algo:ctmq}.

\begin{algorithm}[t]
 \caption{CTMQ: cyclic training with multiple quantization steps for $k$-bit quantized CNNs }
 \small 
 \label{algo:ctmq}
  \SetKwFunction{Loss}{$\mathcal{L}_{q_n}$}
  \SetKwFunction{Return}{Return}
  \SetKwInOut{KwIn}{Input}
  \SetKwInOut{KwOut}{Output}
  \KwIn {initial weights $w^{(0)}$, number of cycles $C$, numbers of iterations $T_s, T_c, T_f$, learning rate $\gamma$, $n$-bit quantizations $q_n$}
  \KwOut {final weights $w^{(T_l)}$}
  \For(\tcp*[h]{1\textsuperscript{st} part: soft knowledge transfer}){$n \leftarrow 8 $ \KwTo $k+2$}  
  {
    \For{$t \leftarrow 0$ \KwTo $T_s - 1$}
    {
      $w^{(t + 1)} = w^{(t)} + \gamma \nabla$ \Loss{$w^{(t)}_{q_n}$}\;
    }
    $w^{(0)} =  w^{(T_s)}$\; 
  }
  \For(\tcp*[h]{2\textsuperscript{nd} part: cyclic training with $q_{k+1}$ and $q_k$}){$c \leftarrow 0 $ \KwTo $C-1$}   {
    \For{$n \leftarrow k+1 $ \KwTo $k$}
    {
      \For{$t \leftarrow 0$ \KwTo $T_c - 1$}
      {
        $w^{(t + 1)} = w^{(t)} + \gamma \nabla$ \Loss{$w^{(t)}_{q_n}$}\;
      }
      $w^{(0)} =  w^{(T_c)}$\; 
    }
  }
  \For{$t \leftarrow 0$ \KwTo $T_c - 1$}
  {
    $w^{(t + 1)} = w^{(t)} + \gamma \nabla$ \Loss{$w^{(t)}_{q_{k+1}}$}\;
  }
  $w^{(0)} =  w^{(T_c)}$\; 
  \For(\tcp*[h]{3\textsuperscript{rd} part: full training with $q_k$}){$t \leftarrow 0$ \KwTo $T_f - 1$} 
  {
    $w^{(t + 1)} = w^{(t)} + \gamma \nabla$ \Loss{$w^{(t)}_{q_k}$}\;
  }
  \Return{$w^{(T_l)}=w^{(T_f)}$}.\; 
\end{algorithm}

In Algorithm~\ref{algo:ctmq}, there are three parts: 
In the first part, after training an $n$-bit quantized model with $T_s$ iterations, 
the pretrained model is used in the initialization of an $(n-1)$-bit quantized model.
The initial weights from the pretrained real-valued model can be used as $w^{(0)}$. 
In $n$-bit quantization, both weights and activations are quantized into $n$ bits.
This iteration commented as \textit{soft knowledge transfer} is repeated until $n=k+2$ (lines 1-6). 
By letting $w^{(0)}$ be $w^{(T_s)}$ (line 5), 
the training results of an accurate model are delivered into an inaccurate model.  
In the second part, cyclic training with $q_{k+1}$ and $q_k$ quantized models are performed during $C$ cycles
(lines 7-18).
Alternatively, $(k+1)$-bit and $k$-bit quantized models are trained during $T_c$ iterations.
The trained weights are used to initialize the next quantized model by $w^{(0)} =  w^{(T_c)}$ (line 12).
Then, a $(k+1)$-bit quantized model is trained after finishing the cyclic training (lines 15-17).
In the third part, 
the trained weights with the final $(k+1)$-bit quantized model is used 
to initialize the final $k$-bit quantized model.
Finally, the $k$-bit quantized model is trained during $T_f$ iterations. 

\subsection{Target quantizations}

In the multiple quantization steps of the proposed method, 
we adopts the quantization scheme shown in DoReFa-Net (\citet{zhou2016dorefa}) 
for quantizing weights and activations, 
which is summarized in the following.
It is noted that the proposed training cannot depend on any specific quantization method. 

Except for binarized weights,
weights $w$ are normalized into $w_{norm}$ for obtaining $k$-bit quantized weights as:

\begin{equation}\label{eq:nbit_norm}
 w_{norm} = \frac{w}{2 * max(abs(tanh(w)))} + 0.5. 
\end{equation}

Terms $tanh$ and $abs$ mean the hyperbolic tangent and absolute functions, respectively.
Notably, $tanh$ function is used to limit the value range. 
Then, normalized weights $w_{norm}$ are quantized into $w_{q_k}$ as:  
\begin{equation}\label{eq:nbit_quant}
 w_{q_k} = 2 \times \frac{round(w_{norm} \times (2^k-1))}{2^k-1} -1. 
\end{equation}

Besides, binarized weights can be obtained using the $sign$ function as: 

\begin{equation}\label{eq:1bit_quant}
  w_{q_1} = \left\{\begin{matrix}
 & sign(w / mean(abs(w))) * mean(abs(w)) \textrm{  if } w  \neq 0, \\ 
 & 1  \textrm{  else.} 
\end{matrix}\right.
\end{equation}
In (\ref{eq:1bit_quant}), when $w=0$, let $sign(w / mean(w))$ be $1$ to avoid the cases with $w_{q_1}= 0$.

On the other hand, $k$-bit quantized activations denoted as $x_{q_k}$
are obtained as:
\begin{equation}\label{eq:nbit_quant_activation}
 x_{q_k} = \frac{round(clamp(x, 0, 1) \times (2^k-1)))}{2^k-1}. 
\end{equation}
In (\ref{eq:nbit_quant_activation}), term $clamp(x, 0, 1)$ denotes the clamp function 
to limit $x$ to the range $[0, 1]$.
Unlike the $1$-bit quantized weights, 
the proposed method does not use the $sign$ function for $1$-bit quantized activations. 
Therefore, $x_{q_1}$ can be $0$ or $1$.

A value $2^k-1$ is multiplied by the denominator and the numerator in (\ref{eq:nbit_quant}) and  
(\ref{eq:nbit_quant_activation}).
In (\ref{eq:1bit_quant}), $mean(abs(w))$ is multiplied by both the denominator and numerator.
The range of values in $k$-bit quantized weights are limited by normalizing weights in (\ref{eq:nbit_norm}). 
In $k$-bit quantized activations, the clamp function in (\ref{eq:nbit_quant_activation}) 
limits the range of values. 
Regardless of $k$, the range of values is fixed so that two quantized models with different $k$ 
have the same range of weights and activations.

\section{Experimental results and analysis}
\label{section:Experiments}
In experiments, existing binarized CNN models were trained  
to evaluate the proposed method on the training of low-bit quantized CNNs.
Binarized CNNs adopted $1$-bit weights and activations in the convolutions 
so that significant computational and storage resources can be saved.   
However, the performance degradation of binarized CNNs has been the most critical issue (\citet{qin2020binary}). 
It is expected that the proposed method can solve this problem by achieving high classification accuracy.
The pyramid structure with highly stacked convolution layers 
 and residual shortcuts have been adopted in many studies of binarized CNNs (\citet{rastegari2016xnor, lin2017towards, 
liu2018bi, bulat2019xnor, phan2020mobinet, liu2020reactnet, kim2021aresb, xu2021improving, shin2022presb}).   
Considering the popularity of the residual binarized networks, 
our experiments adopted a residual CNN model called ResNet-18~\citet{he2016deep}. 
Standardized datasets (CIFAR-100 (\citet{krizhevsky2009learning})) and ImageNet (\citet{ILSVRC15})) 
were in the evaluations.
ResNet-18 was quantized 
for achieving the final trained model with binarized (1-bit) weights and activations. 
For fair comparison, we adopted the quantization scheme in~\citet{zhou2016dorefa}.
Besides, we maintained the structure of basic blocks in DoReFa-Net (\citet{zhou2016dorefa}) and XNOR-Net (\citet{rastegari2016xnor} in the experiments.

\subsection{Experimental frameworks}
We evaluated the proposed training method using
the Pytorch deep learning frameworks (\citet{paszke2019pytorch}).
We coded our training script that supported multiple GPU-based data parallelism.
After finishing training, its real-valued trained model was stored 
in a file with \textit{pth} extension.
As shown in Algorithm~\ref{algo:ctmq}, the file having pretrained weights was loaded for 
the initialization used in the next training.
The above process was repeated in the proposed cyclic training using shell scripts. 
It must be noted that quantizations were not implemented in silicon-based hardware circuits. 
Instead, quantizers were coded as functions 
to emulate the quantizations of weights and activations in the forward path.

\subsection{Target dataset and model structures}

\begin{figure}[!t]
  \centering
  \includegraphics[width=0.6\linewidth]{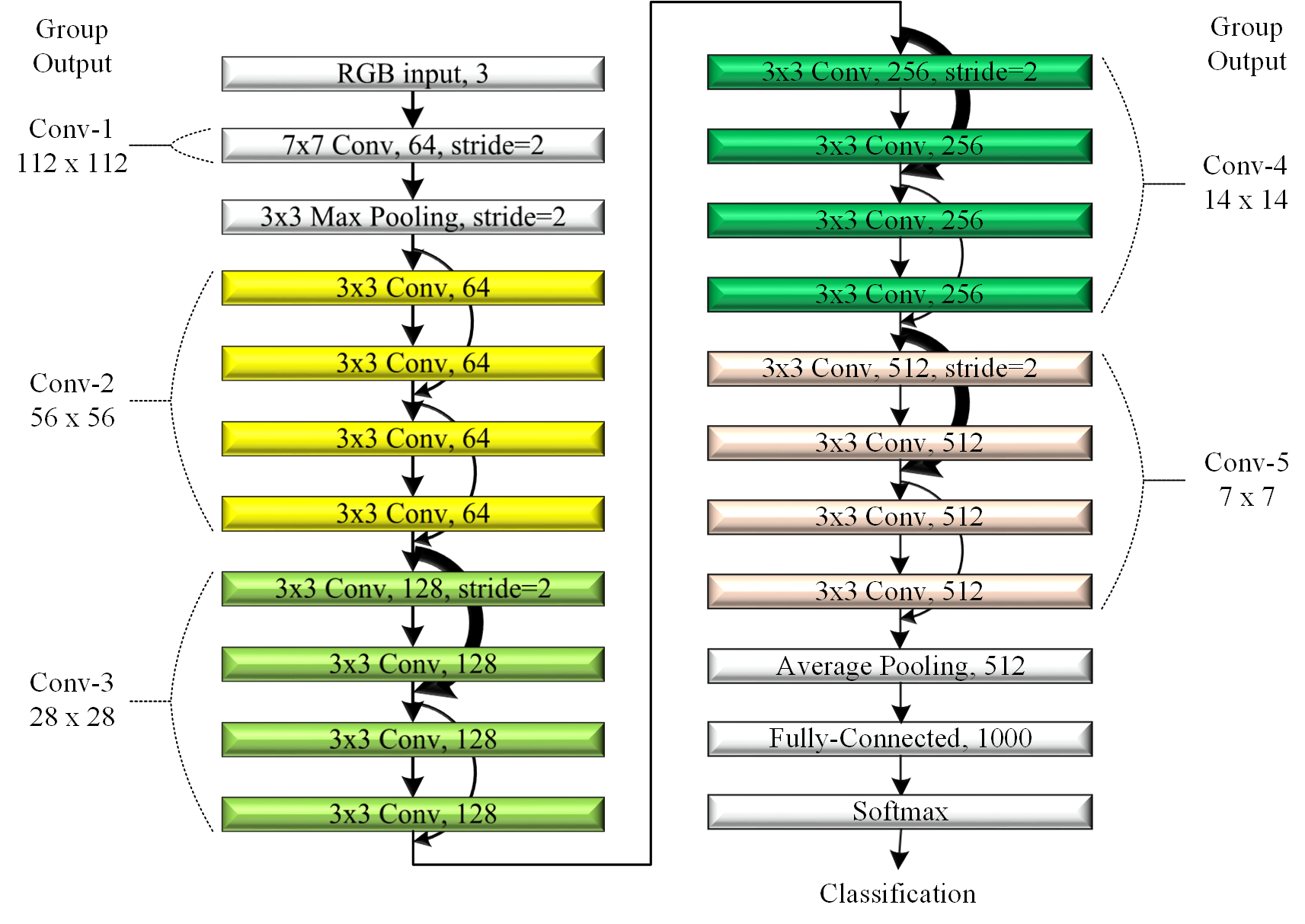}
  \caption
  {
     ResNet-18 for ImageNet dataset.
  }
  \label{fig:resnet18}
\end{figure}

\begin{figure}[!t]
  \centering
  \includegraphics[width=0.8\linewidth]{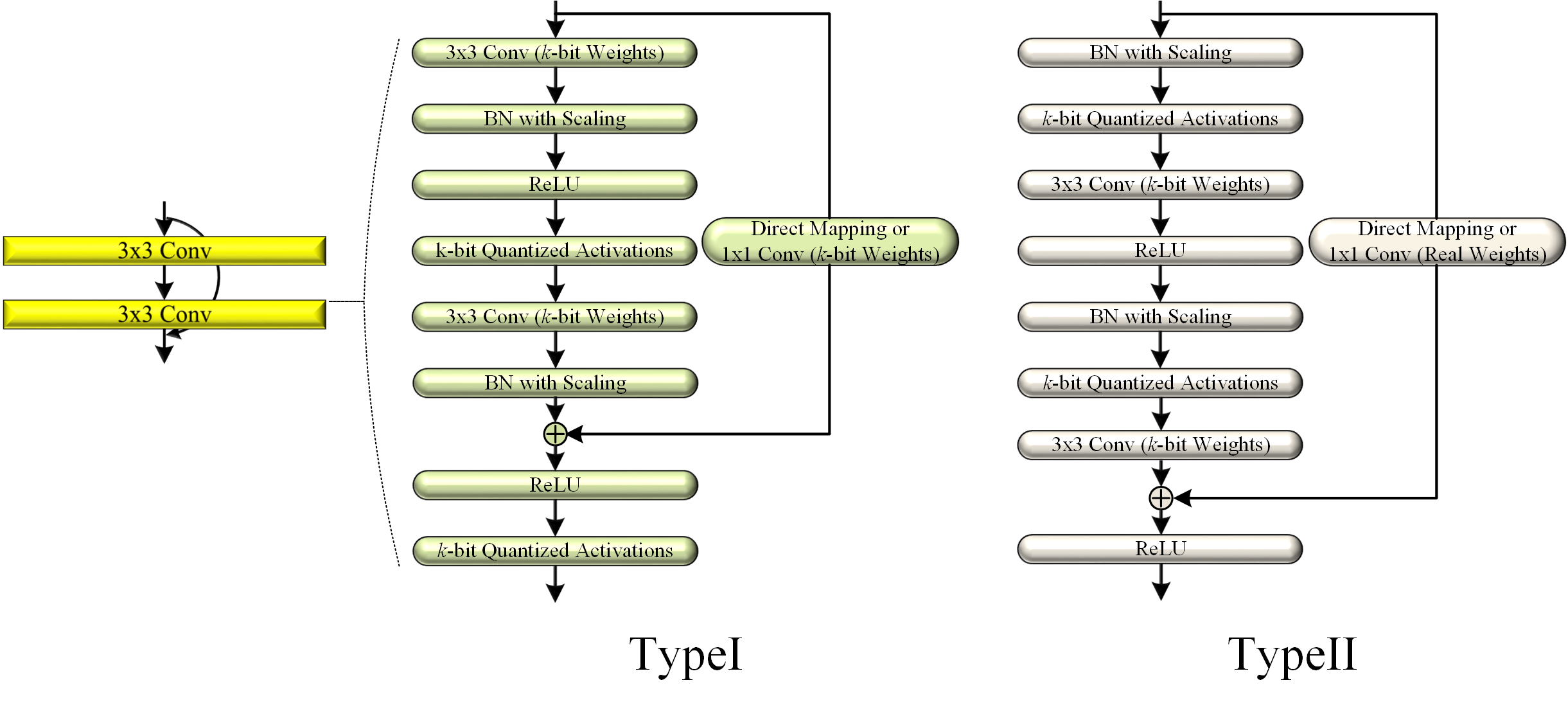}
  \caption
  {
     Basic blocks for quantized ResNet-18: (a) TypeI; (b) TypeII.
  }
  \label{fig:basic_block}
\end{figure}

The summary of target datasets is as follows: 
the CIFAR-100 dataset consists of 60K $32 \times 32$ colour images with 100 classes, 
where each class contains 500 training and 100 test images.
For the data augmentation during training, 
$32 \times 32$ input images were randomly cropped and flipped from $40 \times 40$ padded images.
The ImageNet dataset contains 1.3M training and 50K validation images with 1K classes.   
During training on the ImageNet dataset, images were randomly resized between 256 and 480 pixels, 
and then randomly cropped and flipped into $224 \times 224$ images.
In inference, we adopted $224 \times 224$ center-cropped images without data augmentations
from the validation dataset.

Figure~\ref{fig:resnet18} describes the structure of ResNet-18 for the ImageNet dataset. 
The layers have 5 groups from Conv-1 to Conv-5, where the convolutional layer in a group 
has the same number of output channels. 
Firstly, 3 channels from RGB coloured image are used as $7 \times 7$ convolutional layer with $stride=2$ (Conv-1).
With $stride=2$, the width and height of output features are reduced to half. 
After the max pooling with $stride=2$, the convolutional layers of Conv-2 are performed with 64 input channels.For the CIFAR dataset, a $3 \times 3$ convolutional layer with $stride=1$ is used in Conv-1.
A shortcut with direct mapping are described as a thin round arrow. 
Per two convolutional layers, one shortcut is placed 
so that a basic block contains two convolutional layers and one shortcut.
After finishing two basic blocks, 
the downsampling is performed by doubling the number of channels and having $stride=2$.
The thick round arrow denotes the downsampling using an $1 \times 1$ convolution.
This process is repeated until finishing convolutions in Conv-5. 
Then, the average pooling is applied to 512 channels with $7 \times 7$ features. 
After performing fully-connected and softmax layers, the final classification results can be obtained.
If the quantization is adopted in the first convolution layer of Conv-1, 
the original $8$-bit image pixel data could be degraded.
Because the width and height of features in Conv-5 are small, 
large quantization errors can be critical in the fully-connected layer.
In our experiments,
the convolution in Conv-1 and fully-connected layers adopted real-valued weights and activations 
like \citet{zhou2016dorefa, rastegari2016xnor}.

Figure~\ref{fig:basic_block} illustrates two basic blocks denoted as \textit{TypeI} and \textit{TypeII}.
Our experments designed the TypeI and TypeII basic blocks 
based on DoReFa-Net (\citet{zhou2016dorefa}) and XNOR-Net (\citet{rastegari2016xnor}).
Term BN denotes the batch normalization (\citet{ioffe2015batch}). 
The learnable mean and variance were used to scale and shift the normalized activations.  
In the two convolutional layers and downsampling, $k$-bit weights and activations were used
except for the first convolution in Conv-2.   
Besides, whereas TypeI adopted $k$-bit quantization in the downsampling, 
real-valued weights and activations were used in the downsampling of TypeII. 
In TypeII, after the scaling of the batch normalization layer, $k$-bit quantized activations were obtained,
where the quantization in (\ref{eq:nbit_quant_activation}) was used
instead of the $sign$ function in the original XNOR-Net.

\subsection{Training on CIFAR-100 and ImageNet datasets}
Training on the CIFAR-100 dataset
used ADAM optimizer (\citet{kingma2014adam}) having $\beta_1=0.9 \& \beta_2=0.999$ without weight decay. 
When training a $k$-bit quantized model during $E_{epochs}$ epochs,
the base learning rate $lr_{base}$ was set as 0.001. 
The learning rate $lr$ was decayed based on \textit{poly} policy in the $e_{epochs}$-th epoch, 
limiting the maximum learning rate of the ADAM optimizer by $lr_{base} \times (1- e_{epochs}/E_{epochs})$.
After finishing the training of an $n$-bit quantized model, 
$lr_{base}$ of the next training was reset as 0.001.

The inputs in (\ref{algo:ctmq}) were given as follows:
pretrained real-valued weights were used as the initial weights $w^{(0)}$.
Let $C$ and $q_n$ be 9 and 8, respectively. 
Therefore, an $8$-bit quantized model was firstly trained. 
When finishing the first part in (\ref{algo:ctmq}), a trained $3$-bit quantized model was obtained.
In the second part, $2$-bit and $1$-bit quantized models were alternatively trained during 9 cycles. 
Its mini-batch size was set as 512. 20 training epochs were adopted for $T_s$ and $T_c$ so that $T_s=T_c=1,960$.
The third part peformed 200 training epochs for $T_f=19,600$.   

Several hyperparameters used in the training on the ImageNet dataset were the same 
as those on the CIFAR-100 dataset.
Besides, $C$ and $q_n$ were set as 9 and 8 in the training on the ImageNet dataset.
Considering a large number of images on the ImageNet dataset, 
let $T_s=T_c=12,515$ for 5 training epochs in the first and second parts.
In the third part, 50 training epochs were performed for $T_f=125,150$. 

\begin{figure}[!t]
  \centering
  \includegraphics[width=0.8\linewidth]{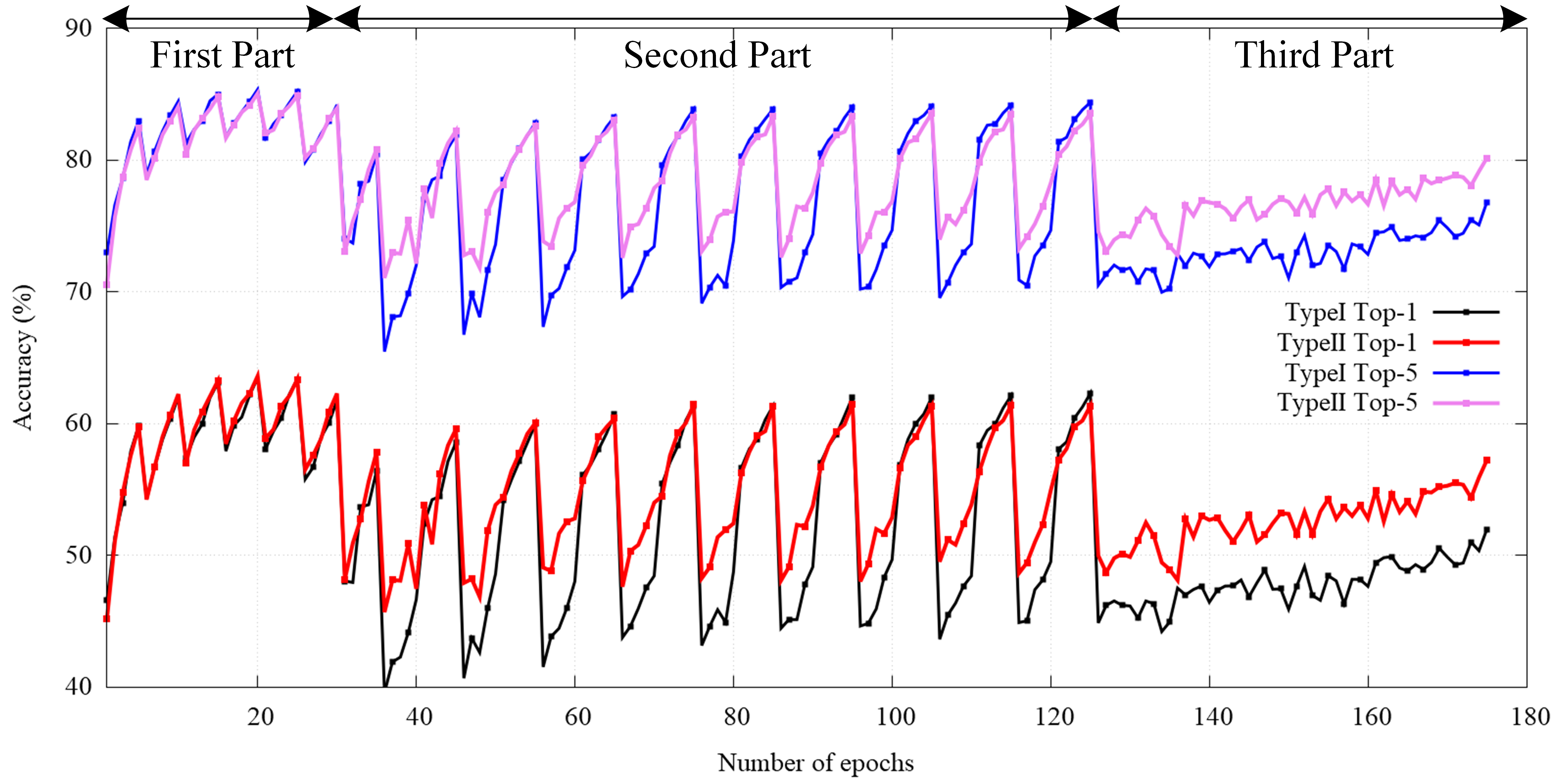}
  \caption
  {
     Inference accuracies of quantized ResNet-18 models on ImageNet dataset.
  }
  \label{fig:accuracies}
\end{figure}

Figure~\ref{fig:accuracies} illustrates inference accuracies of quantized ResNet-18 models 
on the ImageNet dataset during training.
At each epoch, 
the inference accuracies of two binarized ResNet-18 models based on TypeI and TypeII were shown. 
In~\citet{resnet18}, 
it must be noted that 1-crop Top-1 and Top-5 inference accuracies of the original ResNet-18 
were 69.76\% and 89.08\%, respectively.
At the starting point of the first part, 
the quantization error lowered the inference accuracy.
As training continued, the accuracies rapidly increased. 
In the first part, the accuracies from TypeI and TypeII were nearly close,
which means that the effects of the quantized downsampling were not significant in the first part.
In the second part, compared with those of TypeII, 
the models using TypeI had low inference accuracies immediately after $1$-bit quantized models were initialized.
Therefore, it was concluded that the $1$-bit quantization of the downsampling can degrade performance 
in binarized CNNs.  
The initialization of binarized CNNs with pretrained $2$-bit quantized models was performed
at the 36-th, 46-th, and 56-th epochs. 
The inference accuracies at the epochs increased as the cyclic training proceeded. 
For example, the accuracies at the 46-th epoch were better than those at the 36-th epoch
so that better initial weights can be obtained in the cyclic training. 
After the third cycle, the increase slowed down, as shown in Figure~\ref{fig:accuracies}. 
In the third part, compared with the earlier epochs of the second part, 
the $1$-bit quantized CNNs at the third part can start their training
with higher inference accuracies after weight initializations.
As training proceeded, the inference accuracies gradually increased in the third part. 

\begin{table}
  \caption{Comparison of $k$-bit quantized models in terms of inference accuracies}
  \label{table:comparison}
  \centering
  \begin{tabular}{llccccc}
    \toprule
    Dataset                    & Model                  & \textbf{CTMQ} & $k$ & Initialization &Top-1 & Top-5\\
    \midrule
    \midrule
    \multirow{10}{*}{CIFAR-100}& \multirow{5}{*}{TypeI} & N & 32  & N & 73.91\% & -   \\  
                               &                        & N & 8   & N & 68.28\% & -   \\  
                               &                        & N & 1   & N & 64.13\% & -   \\  
                               &                        & N & 1   & Y & 65.41\% & -   \\  
                               &                        & \textbf{Y} & 1   & - & \textbf{67.93\%} & -   \\  
     \cmidrule(r){2-7}
                               & \multirow{5}{*}{TypeII}& N & 32  & N & 72.61\% & -   \\  
                               &                        & N & 8   & N & 67.21\% & -   \\  
                               &                        & N & 1   & N & 67.69\% & -   \\  
                               &                        & N & 1   & Y & 68.89\% & -   \\  
                               &                        & \textbf{Y} & 1 & Y & \textbf{68.84\%} & -   \\  
    \midrule
    \multirow{8}{*}{ImageNet}  & ResNet-18              & N & 32 & N & 69.76\% & 89.08\%   \\  

    \cmidrule(r){2-7}
                               & \multirow{4}{*}{TypeI} & N & 32  & N & 67.40\% & 87.71\% \\  
                               &                        & N & 1   & N & 47.51\% & 72.71\% \\  
                               &                        & N & 1   & Y & 42.38\% & 67.91\% \\  
                               &                        & \textbf{Y} & 1  & Y & \textbf{51.54\%} & \textbf{76.15\%} \\  
    \cmidrule(r){2-7}
                               & TypeII                 & N & 32  & N & 67.03\% & 87.27\% \\  
                               & XNOR-Net               & N & 1 & N & 51.2\% & 73.20\% \\  
                               & TypeII                 & \textbf{Y}  & 1 & N & \textbf{57.0\%} & \textbf{80.05\%} \\  
    \bottomrule
  \end{tabular}
\end{table}

Table~\ref{table:comparison} lists the comparison in terms of inference accuracies. 
The proposed training method was evaluated based on the average of five runs. 
The accuracies of ResNet-18 and XNOR-Net in Table~\ref{table:comparison} 
were referenced from~\citet{resnet18} and~\citet{rastegari2016xnor}.
There were no published results on the quantized ResNet-18 on the CIFAR-100 dataset 
in~\citet{zhou2016dorefa} and~\citet{rastegari2016xnor}
so that we trained the quantized models during 200 epochs on the CIFAR-100 dataset 
and 50 epochs on the ImageNet dataset for the counterparts without applying the proposed method.
In the training of the counterparts, 
most of the hyperparameters were identical with those using the proposed method.
However, weight decay was set as 0.0001 for the regularization. 
In Table~\ref{table:comparison}, when the initialization was Y, 
real-valued pretrained models were used in weight initialization. 
Besides, when item CTMQ was N, the soft knowledge transfer and cyclic training were not applied.

Due to the small number of training images and small image size in the CIFAR-100 dataset,
overfitting was shown in the experiments of ResNet-18.
Therefore, the enhancements with the proposed method were limited in the experiments on the CIFAR-100 dataset.
On the other hand, the proposed method significantly enhanced the final inference accuracies on the ImageNet dataset 
compared with other cases.
On the ImageNet dataset, when not using the proposed method, TypeI with initialization
showed significant performance drop only having 42.38\% Top-1 and 67.91\% Top-5 inference accuracies.
TypeI using the proposed CTMQ achieved 51.54\% Top-1 and 76.15\% Top-5 accuracies, 
which produced better performance over the original XNOR-Net~\cite{rastegari2016xnor}.
The binarized ResNet-18 with TypeII 
had Top-1 and Top-5 inference accuracies up to 57.0\% and 80.05\%, respectively,
which showed 5.80\% and 6.85\% enhancements over the training results of the original XNOR-Net.

Considering the experimental results, 
it is concluded that the proposed training method was effective for training complex residual networks
on large dataset. 

\section{Conclusion}
\label{section:Conclusion}
We focus on a new training method performing cyclic training with different $k$-bit quantizations.
The training method performs multiple cyclic training 
for softly delivering the knowledge of pretrained accurate models 
and using better training ability of accurate networks.
Although the number of training epochs increases during three training parts, 
it has been proven that the final inference accuracies for complex residual networks on large dataset 
are significantly enhanced in experimental results. 
It is noted that there are no structural changes in CNNs for enhancing the final inference accuracy. 
Although several binarized CNNs are trained in experiments, 
the proposed method can be applied to other quantized CNNs based on the existing quantization and optimizer. 
Notably, the training method can advance Top-1 and Top-5 accuracies of the binarized ResNet-18
on the ImageNet dataset by 5.80\% and 6.85\%, respectively.
Considering the enhanced classification outputs, 
it is concluded that the proposed cyclic training method is useful
for producing high-performance systems in quantized CNNs.

\newpage

\section*{References}

\medskip

\bibliographystyle{plainnat}
{
\small
\bibliography{nips2022}

\begin{thebibliography}{38}
\providecommand{\natexlab}[1]{#1}
\providecommand{\url}[1]{\texttt{#1}}
\expandafter\ifx\csname urlstyle\endcsname\relax
  \providecommand{\doi}[1]{doi: #1}\else
  \providecommand{\doi}{doi: \begingroup \urlstyle{rm}\Url}\fi

\bibitem[Bengio et~al.(2013)Bengio, L{\'e}onard, and
  Courville]{bengio2013estimating}
Yoshua Bengio, Nicholas L{\'e}onard, and Aaron Courville.
\newblock Estimating or propagating gradients through stochastic neurons for
  conditional computation.
\newblock \emph{arXiv preprint arXiv:1308.3432}, 2013.

\bibitem[Bulat and Tzimiropoulos(2019)]{bulat2019xnor}
Adrian Bulat and Georgios Tzimiropoulos.
\newblock Xnor-net++: Improved binary neural networks.
\newblock \emph{arXiv preprint arXiv:1909.13863}, 2019.

\bibitem[Courbariaux et~al.(2016)Courbariaux, Hubara, Soudry, El-Yaniv, and
  Bengio]{courbariaux2016binarized}
Matthieu Courbariaux, Itay Hubara, Daniel Soudry, Ran El-Yaniv, and Yoshua
  Bengio.
\newblock Binarized neural networks: Training deep neural networks with weights
  and activations constrained to+ 1 or-1.
\newblock \emph{arXiv preprint arXiv:1602.02830}, 2016.

\bibitem[Deng et~al.(2017)Deng, Jiao, Pei, Wu, and Li]{deng2017gated}
Lei Deng, Peng Jiao, Jing Pei, Zhenzhi Wu, and Guoqi Li.
\newblock Gated xnor networks: deep neural networks with ternary weights and
  activations under a unified discretization framework.
\newblock \emph{arXiv preprint arXiv:1705.09283}, 2, 2017.

\bibitem[Gholami et~al.(2021)Gholami, Kim, Dong, Yao, Mahoney, and
  Keutzer]{gholami2021survey}
Amir Gholami, Sehoon Kim, Zhen Dong, Zhewei Yao, Michael~W Mahoney, and Kurt
  Keutzer.
\newblock A survey of quantization methods for efficient neural network
  inference.
\newblock \emph{arXiv preprint arXiv:2103.13630}, 2021.

\bibitem[Gupta et~al.(2015)Gupta, Agrawal, Gopalakrishnan, and
  Narayanan]{gupta2015deep}
Suyog Gupta, Ankur Agrawal, Kailash Gopalakrishnan, and Pritish Narayanan.
\newblock Deep learning with limited numerical precision.
\newblock In \emph{International Conference on Machine Learning}, pages
  1737--1746, 2015.

\bibitem[Han et~al.(2015)Han, Mao, and Dally]{han2015deep}
Song Han, Huizi Mao, and William~J Dally.
\newblock Deep compression: Compressing deep neural networks with pruning,
  trained quantization and huffman coding.
\newblock \emph{arXiv preprint arXiv:1510.00149}, 2015.

\bibitem[Hassani(2009)]{hassani2009dirac}
Sadri Hassani.
\newblock Dirac delta function.
\newblock In \emph{Mathematical methods}, pages 139--170. Springer, 2009.

\bibitem[He et~al.(2016)He, Zhang, Ren, and Sun]{he2016deep}
Kaiming He, Xiangyu Zhang, Shaoqing Ren, and Jian Sun.
\newblock Deep residual learning for image recognition.
\newblock In \emph{Proceedings of the IEEE conference on computer vision and
  pattern recognition}, pages 770--778, 2016.

\bibitem[Helwegen et~al.(2019)Helwegen, Widdicombe, Geiger, Liu, Cheng, and
  Nusselder]{helwegen2019latent}
Koen Helwegen, James Widdicombe, Lukas Geiger, Zechun Liu, Kwang-Ting Cheng,
  and Roeland Nusselder.
\newblock Latent weights do not exist: Rethinking binarized neural network
  optimization.
\newblock In \emph{Advances in neural information processing systems}, pages
  7531--7542, 2019.

\bibitem[Hubara et~al.(2016)Hubara, Courbariaux, Soudry, El-Yaniv, and
  Bengio]{hubara2016binarized}
Itay Hubara, Matthieu Courbariaux, Daniel Soudry, Ran El-Yaniv, and Yoshua
  Bengio.
\newblock Binarized neural networks.
\newblock In \emph{Advances in neural information processing systems}, pages
  4107--4115, 2016.

\bibitem[Hubara et~al.(2017)Hubara, Courbariaux, Soudry, El-Yaniv, and
  Bengio]{hubara2017quantized}
Itay Hubara, Matthieu Courbariaux, Daniel Soudry, Ran El-Yaniv, and Yoshua
  Bengio.
\newblock Quantized neural networks: Training neural networks with low
  precision weights and activations.
\newblock \emph{The Journal of Machine Learning Research}, 18\penalty0
  (1):\penalty0 6869--6898, 2017.

\bibitem[Ioffe and Szegedy(2015)]{ioffe2015batch}
Sergey Ioffe and Christian Szegedy.
\newblock Batch normalization: Accelerating deep network training by reducing
  internal covariate shift.
\newblock \emph{arXiv preprint arXiv:1502.03167}, 2015.

\bibitem[Kim et~al.(2020)Kim, Kim, Kim, and Kim]{kim2020binaryduo}
Hyungjun Kim, Kyungsu Kim, Jinseok Kim, and Jae-Joon Kim.
\newblock Binaryduo: Reducing gradient mismatch in binary activation network by
  coupling binary activations.
\newblock \emph{arXiv preprint arXiv:2002.06517}, 2020.

\bibitem[Kim(2021)]{kim2021aresb}
HyunJin Kim.
\newblock Aresb-net: accurate residual binarized neural networks using shortcut
  concatenation and shuffled grouped convolution.
\newblock \emph{PeerJ Computer Science}, 7:\penalty0 e454, 2021.

\bibitem[Kingma and Ba(2014)]{kingma2014adam}
Diederik~P Kingma and Jimmy Ba.
\newblock Adam: A method for stochastic optimization.
\newblock \emph{arXiv preprint arXiv:1412.6980}, 2014.

\bibitem[Krizhevsky et~al.(2009)Krizhevsky, Hinton,
  et~al.]{krizhevsky2009learning}
Alex Krizhevsky, Geoffrey Hinton, et~al.
\newblock Learning multiple layers of features from tiny images.
\newblock 2009.

\bibitem[Lin et~al.(2017)Lin, Zhao, and Pan]{lin2017towards}
Xiaofan Lin, Cong Zhao, and Wei Pan.
\newblock Towards accurate binary convolutional neural network.
\newblock In \emph{Advances in Neural Information Processing Systems}, pages
  345--353, 2017.

\bibitem[Liu et~al.(2018)Liu, Wu, Luo, Yang, Liu, and Cheng]{liu2018bi}
Zechun Liu, Baoyuan Wu, Wenhan Luo, Xin Yang, Wei Liu, and Kwang-Ting Cheng.
\newblock Bi-real net: Enhancing the performance of 1-bit cnns with improved
  representational capability and advanced training algorithm.
\newblock In \emph{Proceedings of the European conference on computer vision
  (ECCV)}, pages 722--737, 2018.

\bibitem[Liu et~al.(2020)Liu, Shen, Savvides, and Cheng]{liu2020reactnet}
Zechun Liu, Zhiqiang Shen, Marios Savvides, and Kwang-Ting Cheng.
\newblock Reactnet: Towards precise binary neural network with generalized
  activation functions.
\newblock \emph{arXiv preprint arXiv:2003.03488}, 2020.

\bibitem[Martinez et~al.(2020)Martinez, Yang, Bulat, and
  Tzimiropoulos]{martinez2020training}
Brais Martinez, Jing Yang, Adrian Bulat, and Georgios Tzimiropoulos.
\newblock Training binary neural networks with real-to-binary convolutions.
\newblock \emph{arXiv preprint arXiv:2003.11535}, 2020.

\bibitem[Paszke et~al.(2019)Paszke, Gross, Massa, Lerer, Bradbury, Chanan,
  Killeen, Lin, Gimelshein, Antiga, et~al.]{paszke2019pytorch}
Adam Paszke, Sam Gross, Francisco Massa, Adam Lerer, James Bradbury, Gregory
  Chanan, Trevor Killeen, Zeming Lin, Natalia Gimelshein, Luca Antiga, et~al.
\newblock Pytorch: An imperative style, high-performance deep learning library.
\newblock In \emph{Advances in Neural Information Processing Systems}, pages
  8024--8035, 2019.

\bibitem[Phan et~al.(2020)Phan, He, Savvides, Shen, et~al.]{phan2020mobinet}
Hai Phan, Yihui He, Marios Savvides, Zhiqiang Shen, et~al.
\newblock Mobinet: A mobile binary network for image classification.
\newblock In \emph{The IEEE Winter Conference on Applications of Computer
  Vision}, pages 3453--3462, 2020.

\bibitem[Qin et~al.(2020)Qin, Gong, Liu, Bai, Song, and Sebe]{qin2020binary}
Haotong Qin, Ruihao Gong, Xianglong Liu, Xiao Bai, Jingkuan Song, and Nicu
  Sebe.
\newblock Binary neural networks: A survey.
\newblock \emph{Pattern Recognition}, 105:\penalty0 107281, 2020.

\bibitem[Rastegari et~al.(2016)Rastegari, Ordonez, Redmon, and
  Farhadi]{rastegari2016xnor}
Mohammad Rastegari, Vicente Ordonez, Joseph Redmon, and Ali Farhadi.
\newblock Xnor-net: Imagenet classification using binary convolutional neural
  networks.
\newblock In \emph{European conference on computer vision}, pages 525--542.
  Springer, 2016.

\bibitem[Ruder(2016)]{ruder2016overview}
Sebastian Ruder.
\newblock An overview of gradient descent optimization algorithms.
\newblock \emph{arXiv preprint arXiv:1609.04747}, 2016.

\bibitem[Russakovsky et~al.(2015)Russakovsky, Deng, Su, Krause, Satheesh, Ma,
  Huang, Karpathy, Khosla, Bernstein, Berg, and Fei-Fei]{ILSVRC15}
Olga Russakovsky, Jia Deng, Hao Su, Jonathan Krause, Sanjeev Satheesh, Sean Ma,
  Zhiheng Huang, Andrej Karpathy, Aditya Khosla, Michael Bernstein,
  Alexander~C. Berg, and Li~Fei-Fei.
\newblock {ImageNet Large Scale Visual Recognition Challenge}.
\newblock \emph{International Journal of Computer Vision (IJCV)}, 115\penalty0
  (3):\penalty0 211--252, 2015.
\newblock \doi{10.1007/s11263-015-0816-y}.

\bibitem[Shin and Kim(2022)]{shin2022presb}
Jungwoo Shin and HyunJin Kim.
\newblock Presb-net: parametric binarized neural network with learnable
  activations and shuffled grouped convolution.
\newblock \emph{PeerJ Computer Science}, 8:\penalty0 e842, 2022.

\bibitem[Torch(2022)]{resnet18}
Contributors Torch.
\newblock {Models and Pre-trained weights}.
\newblock \url{https://pytorch.org/vision/stable/models.html}, 2022.
\newblock [Online; accessed 17-May-2022].

\bibitem[Wan et~al.(2018)Wan, Shen, Liu, Zhu, Qin, Shao, and
  Tao~Shen]{wan2018tbn}
Diwen Wan, Fumin Shen, Li~Liu, Fan Zhu, Jie Qin, Ling Shao, and Heng Tao~Shen.
\newblock Tbn: Convolutional neural network with ternary inputs and binary
  weights.
\newblock In \emph{Proceedings of the European Conference on Computer Vision
  (ECCV)}, pages 315--332, 2018.

\bibitem[Wang et~al.(2018)Wang, Choi, Brand, Chen, and
  Gopalakrishnan]{wang2018training}
Naigang Wang, Jungwook Choi, Daniel Brand, Chia-Yu Chen, and Kailash
  Gopalakrishnan.
\newblock Training deep neural networks with 8-bit floating point numbers.
\newblock In \emph{Advances in neural information processing systems}, pages
  7675--7684, 2018.

\bibitem[Wu et~al.(2020)Wu, Judd, Zhang, Isaev, and
  Micikevicius]{wu2020integer}
Hao Wu, Patrick Judd, Xiaojie Zhang, Mikhail Isaev, and Paulius Micikevicius.
\newblock Integer quantization for deep learning inference: Principles and
  empirical evaluation.
\newblock \emph{arXiv preprint arXiv:2004.09602}, 2020.

\bibitem[Wu et~al.(2016)Wu, Leng, Wang, Hu, and Cheng]{wu2016quantized}
Jiaxiang Wu, Cong Leng, Yuhang Wang, Qinghao Hu, and Jian Cheng.
\newblock Quantized convolutional neural networks for mobile devices.
\newblock In \emph{Proceedings of the IEEE Conference on Computer Vision and
  Pattern Recognition}, pages 4820--4828, 2016.

\bibitem[Xu et~al.(2021{\natexlab{a}})Xu, Chen, He, Wang, and
  Cheng]{xu2021improving}
Weixiang Xu, Qiang Chen, Xiangyu He, Peisong Wang, and Jian Cheng.
\newblock Improving binary neural networks through fully utilizing latent
  weights.
\newblock \emph{arXiv preprint arXiv:2110.05850}, 2021{\natexlab{a}}.

\bibitem[Xu et~al.(2021{\natexlab{b}})Xu, Lin, Liu, Chen, Shao, Gao, Tian, and
  Ji]{xu2021recu}
Zihan Xu, Mingbao Lin, Jianzhuang Liu, Jie Chen, Ling Shao, Yue Gao, Yonghong
  Tian, and Rongrong Ji.
\newblock Recu: Reviving the dead weights in binary neural networks.
\newblock \emph{arXiv preprint arXiv:2103.12369}, 2021{\natexlab{b}}.

\bibitem[Yin et~al.(2019)Yin, Lyu, Zhang, Osher, Qi, and
  Xin]{yin2019understanding}
Penghang Yin, Jiancheng Lyu, Shuai Zhang, Stanley Osher, Yingyong Qi, and Jack
  Xin.
\newblock Understanding straight-through estimator in training activation
  quantized neural nets.
\newblock \emph{arXiv preprint arXiv:1903.05662}, 2019.

\bibitem[Yu et~al.(2020)Yu, Wen, Cheng, Sun, Han, and Shi]{yu2020low}
Haibao Yu, Tuopu Wen, Guangliang Cheng, Jiankai Sun, Qi~Han, and Jianping Shi.
\newblock Low-bit quantization needs good distribution.
\newblock In \emph{Proceedings of the IEEE/CVF Conference on Computer Vision
  and Pattern Recognition Workshops}, pages 680--681, 2020.

\bibitem[Zhou et~al.(2016)Zhou, Wu, Ni, Zhou, Wen, and Zou]{zhou2016dorefa}
Shuchang Zhou, Yuxin Wu, Zekun Ni, Xinyu Zhou, He~Wen, and Yuheng Zou.
\newblock Dorefa-net: Training low bitwidth convolutional neural networks with
  low bitwidth gradients.
\newblock \emph{arXiv preprint arXiv:1606.06160}, 2016.

\end{thebibliography}
}
%\section{Appendix}

%Optionally include extra information (complete proofs, additional experiments and plots) in the appendix.
%This section will often be part of the supplemental material.

\end{document}